%% file: arxiv.tex
\documentclass{article}
\usepackage[T2A]{fontenc}
\usepackage[utf8]{inputenc}
\usepackage[french,russian,english]{babel}
\usepackage{amsmath,amssymb}
\usepackage{multirow}
\usepackage{array}
\usepackage{graphicx}
\usepackage{url}

\title{Rotations and Interpretability of Word Embeddings: the Case of the Russian Language}

\author{Alexey Zobnin \\
National Research University Higher School of Economics,\\
Faculty of Computer Science,\\
azobnin@hse.ru}

\date{}

\DeclareMathOperator{\interp}{interp}
\DeclareMathOperator{\tr}{tr}

\begin{document}

\maketitle

\begin{abstract}
Consider a continuous word embedding model.
Usually, the cosines between word vectors are used as a measure of similarity of words.
These cosines do not change under orthogonal transformations of the embedding space.
We demonstrate that, using some canonical orthogonal transformations from SVD,
it is possible both to increase the meaning of some components and to make the components more stable under re-learning.
We study the interpretability of components for publicly available models for the Russian language (RusVect\=or\=es, fastText, RDT).
\end{abstract}

\section{Introduction}


Word embeddings are frequently used in NLP tasks.
In vector space models every word from the source corpus is represented by a dense vector in~$\mathbb{R}^d$,
where the typical dimension~$d$ varies from tens to hundreds.
Such embedding maps similar (in some sense) words to close vectors.
These models are based on the so called distributional hypothesis: similar words tend to occur in similar contexts~\cite{harris1954distributional}.
Some models also use letter trigrams or additional word properties such as morphological tags.

There are two basic approaches to the construction of word embeddings.
The first is count-based, or explicit~\cite{levy2014linguistic,dhillon2015eigenwords}.
For every word-context pair some measure of their proximity (such as frequency or PMI) is calculated.
Thus, every word obtains a sparse vector of high dimension.
Further, the dimension is reduced using singular value decomposition (SVD) or non-negative sparse embedding (NNSE).
It was shown that truncated SVD or NNSE captures latent meaning in such models~\cite{landauer1997solution,murphy2012learning}.
That is why the components of embeddings in such models are already in some sense canonical.
The second approach is predict-based, or implicit.
Here the embeddings are constructed by a neural network.
Popular models of this kind include
word2vec~\cite{mikolov2013efficient,mikolov2013distributed}
and fastText~\cite{bojanowski2016enriching}.

Consider a predict-based word embedding model.
Usually in such models two kinds of vectors, both for words and contexts, are constructed.
Let $N$ be the vocabulary size and $d$ be the dimension of embeddings.
Let $W$ and $C$ be $N \times d$-matrices whose rows are word and context vectors.
As a rule, the objectives of such models depend on the dot products of word and context vectors,
i.~e., on the elements of $WC^T$.
In some models the optimization can be directly rewritten as a matrix factorization problem~\cite{levy2014neural,cotterell2017explaining}.
This matrix remains unchanged under substitutions
$W \mapsto W S, \quad C \mapsto C {S^{-1}}^T$
for any invertible~$S$.
Thus, when no other constraints are specified, there are infinitely many equivalent solutions~\cite{fonarev2017riemannian}.

Choosing a good, not necessarily orthogonal, post-processing transformation~$S$
that improves quality in applied problems is itself interesting enough~\cite{mu2017all}.
However, only word vectors are typically used in practice, and context vectors are ignored.
The cosine distance between word vectors is used as a similarity measure between words.
These cosines will not change if and only if the transformation $S$
is orthogonal.
Such transformations do not affect the quality of the model, but may elucidate the meaning of vectors' components.
Thus, the following problem arises:
\emph{what orthogonal transformation is the best one for describing the meaning of some (or all) components?}

It is believed that the meaning of the components of word vectors is hidden~\cite{gladkova2016intrinsic}.
But even if we determine the ``meaning'' of some component, we may loose it after re-training because of random initialization, thread synchronization issues, etc.
Many researchers~\cite{luo2015online,ruseti2016using,andrews2016compressing,jang2017elucidating} ignore this fact and, say, work with vector components directly,
and only some of them take basis rotations into account~\cite{tsvetkov2016correlation}.
We show that, generally, re-trained model differ from the source model by almost orthogonal transformation.
This leads us to the following problem:
\emph{how one can choose the canonical coordinates for embeddings that are (almost) invariant with respect to re-training?}

We suggest using well-known plain old technique, namely, the singular value decomposition of the word matrix~$W$.
We study the principal components of different models for Russian language (RusVect\=or\=es, RDT, fastText, etc.),
although the results are applicable for any language as well.

\section{Related Work}

Interpretability of the components have been extensively studied for topic models.
In~\cite{chang2009reading,lau2014machine} two methods for estimating the coherence of topic models with manual tagging have been proposed: namely,
word intrusion and topic intrusion.
Automatic measures of coherence based on different similarities of words were proposed in~\cite{aletras2013evaluating,nikolenko2016topic}.
But unlike topic models,
these methods cannot be applied directly to word vectors.


There are lots of new models where interpretability is either taken into account by design
\cite{luo2015online} (modified skip-gram that produces non-negative entries),
or is obtained automagically
\cite{andrews2016compressing} (sparse autoencoding).

Lots of authors try to extract some predefined significant properties from vectors:
\cite{jang2017elucidating} (for non-negative sparse embeddings),
\cite{tsvetkov2016correlation} (using a CCA-based alignment between word vectors and manually-annotated linguistic resource),
\cite{rothe2016word} (ultradense projections).

Singular vector decomposition is the core of count-based models.
To our knowledge, the only paper where SVD was applied to predict-based word embedding matrices is~\cite{mu2017all}.
In~\cite{arora2017simple} the first principal component is constructed for sentence embedding matrix (this component is excluded as the common one).

Word embeddings for Russian language were studied in~\cite{kutuzov2015texts,Kutuzov2015,panchenko2015russe,arefyev2015evaluating}.

\section{Theoretical Considerations}

\subsection{Singular value decomposition}
Let $m \ge n$. Recall~\cite{jolliffe2002principal} that a singular value decomposition (SVD) of an $m\times n$-matrix $M$ is a decomposition $M = U \Sigma V^T$,
where $U$ is an an $m \times n$ matrix, $U^T U = I_{n}$, $\Sigma$ is a diagonal $n \times n$-matrix,
and $V$ is an $n \times n$ orthogonal matrix.
Diagonal elements of $\Sigma$ are non-negative and are called singular values.
Columns of $U$ are eigenvectors of $M M^T$, and columns of $V$ are eigenvectors of $M^T M$.
Squares of singular values are eigenvalues of these matrices.
If all singular values are different and positive,
then SVD is unique up to permutation of singular values and choosing the direction of singular vectors.
Buf if some singular values coincide or equal zero, new degrees of freedom arise.

\subsection{Invariance under re-training}
\newcommand{\Wa}{M_1}
\newcommand{\Wb}{M_2}
\newcommand{\Wc}{M_i}
Learning methods are usually not deterministic.
The model re-trained with similar hyperparameters may have completely different components.
Let $\Wa$ and $\Wb$ be the word matrices obtained after two separate trainings of the model.
Let these embeddings be similar in the sense that cosine distances between words are almost the same,
i.~e., $\Wa \Wa^T \approx \Wb \Wb^T$.
Suppose also that singular values of each $\Wc$ are different and non-zero.
Then one can show that $\Wa$ and $\Wb$ differ only by the (almost) orthogonal factor.
Indeed, left singular vectors in SVD of $\Wc$
are eigenvectors of $\Wc \Wc^T$.
Hence, matrices $U$ and $\Sigma$ in SVD of $\Wa$ and $\Wb$ can be chosen the same.
Thus, $\Wb \approx \Wa Q$, where $Q Q^T = I_d$. Here $Q$ can be chosen as $V_1 V_2^T$
where $V_i$ are matrices of right singular vectors in SVD of $\Wc$.

\subsection{Interpretability measures}
One of traditional measures of interpretability in topic modeling looks as follows~\cite{newman2010automatic,lau2014machine}.
For each component, $n$ most probable words are selected.
Then for each pair of selected words some co-occurrence measure such as PMI is calculated.
These values are averaged over all pairs of selected words and all components.
The other approaches use human markup.
Such measures need additional data, and it is difficult to study them algebraically.
Also, unlike topic modeling, word embeddings are not probabilistic:
both positive and negative values of coordinates should be considered.

Let all word vectors be normalized and $W$ be the word matrix.
Inspired by~\cite{nikolenko2016topic}, where vector space models are used for evaluating topic coherence,
we suggest to estimate the interpretability of $k$th component as
\begin{equation*}
 \interp_k W = \sum_{i,j=1}^N W_{i,k} W_{j,k} \left(W_i \cdot W_j \right).
\end{equation*}
The factors $W_{i,k}$ and $W_{j,k}$ are the values of $k$th components of $i$th and $j$th words.
The dot product $\left(W_i \cdot W_j\right)$ reflects the similarity of words.
Thus, this measure will be high if similar words have similar values of $k$th coordinates.

What orthogonal transformation $Q$ maximizes this interpretability (for some, or all components) of $WQ$?
In matrix terms,
$$
\interp_k W =(W^T W W^T W)_{k, k},
$$
and
$$
 \interp_k WQ = \left(Q^T W^T W W^T W Q\right)_{k,k}
$$
because $Q$ is orthogonal.
The total interpretability over all components is
\begin{gather*}
 \sum_{k=1}^d \interp_k WQ = \sum_{k=1}^d \left(Q^T W^T W W^T W Q \right)_{k,k} = \\
= \tr Q^T W^T W W^T W Q = \tr \left(W^T W W^T W\right) = \sum_{k=1}^d \interp_k W,
\end{gather*}
because
$\tr Q^T X Q = \tr Q^{-1} X Q = \tr X$.
It turns out that \emph{in average} the interpretability is constant under any orthogonal transformation.
But it is possible to make the first components more interpretable due to the other components.
For example,
$$
 (Q^T W^T W W^T W Q)_{1, 1} = \left(q^T W^T W q\right)^2
$$
is maximized when $q$ is the eigenvector of $W^T W$ with the largest singular value, i.~e., the first right singular vector of~$W$~\cite{jolliffe2002principal}.
Let's fix this vector and choose other vectors to be orthogonal to the selected ones and to maximize the interpretability.
We arrive at $Q = V$, where $V$ is the right orthogonal factor in SVD $W = U \Sigma V^T$.


\section{Experiments}

\begin{figure}[!t]
\label{singular_values}
\caption{Decreasing of singular values for the rotated fastText models (dim=100).}
\begin{center}
\includegraphics[width=12cm]{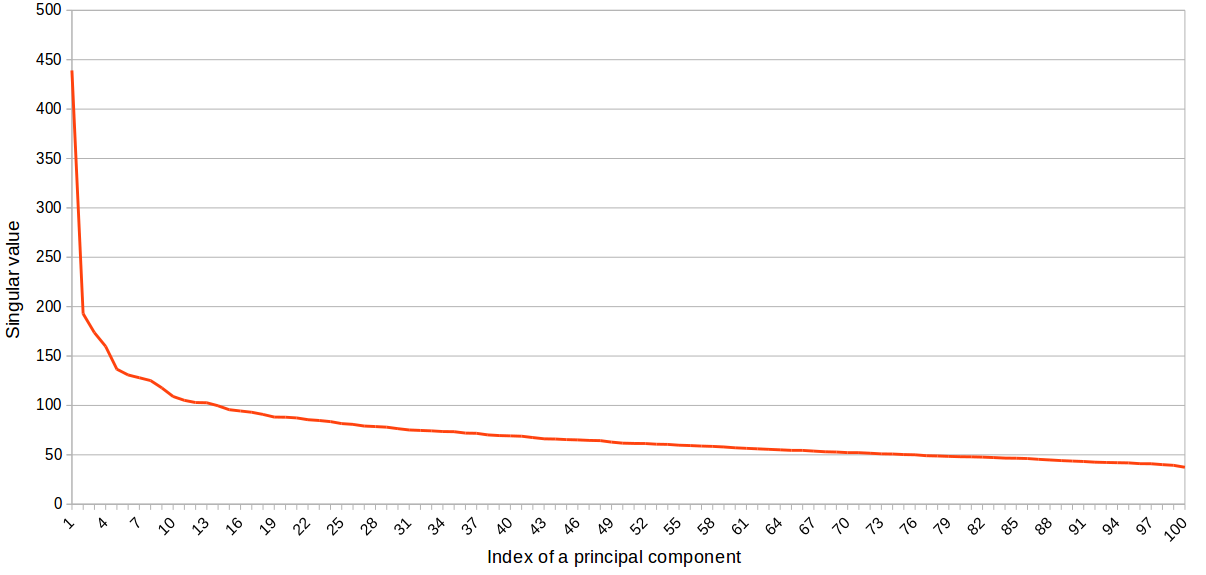}
\end{center}
\end{figure}

\begin{figure}[!t]
\label{overlapping}
\caption{The amount of common top and bottom words for the source models (blue) and the rotated models (red).}
\begin{center}
\includegraphics[width=12cm]{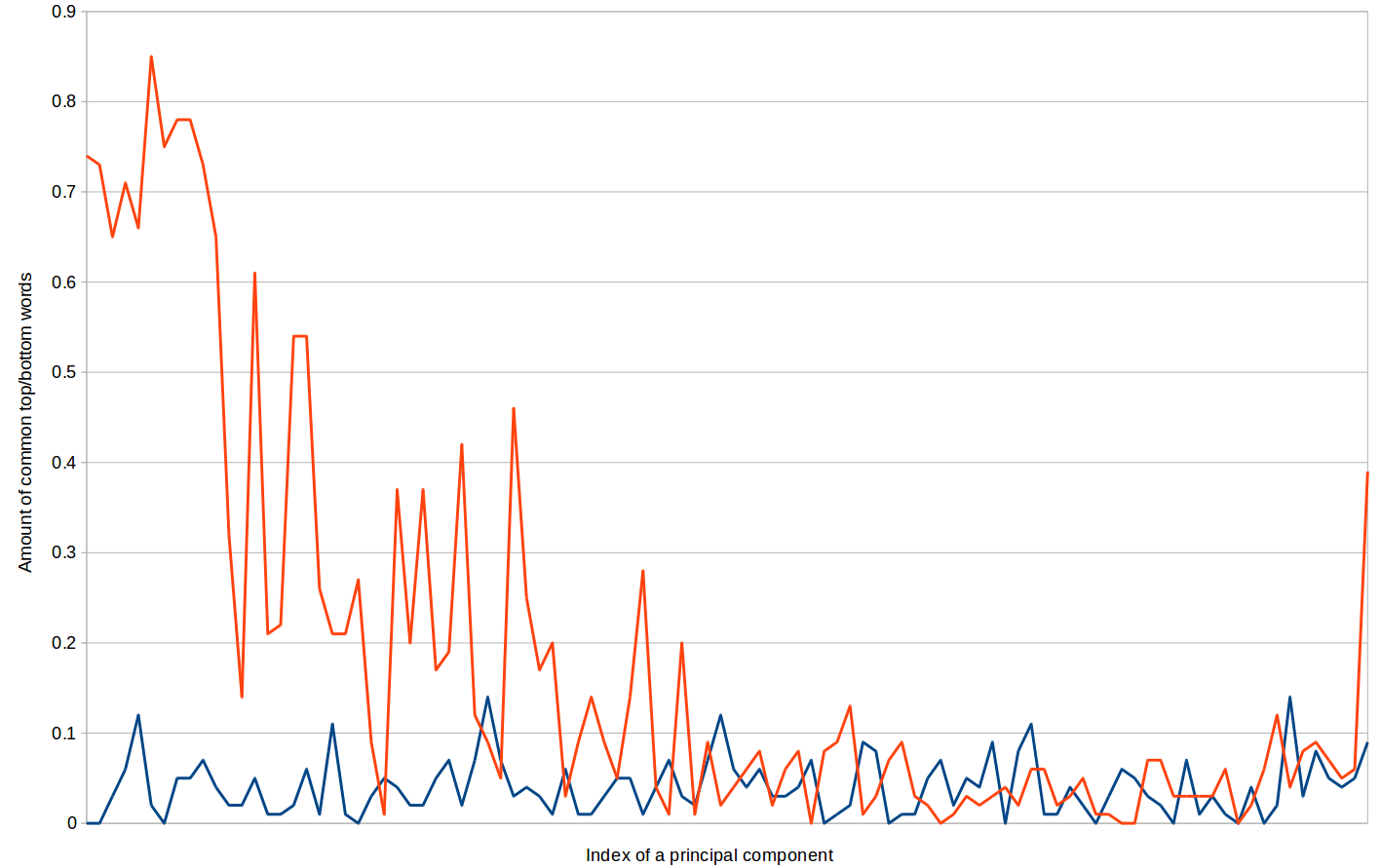}
\end{center}
\label{alignment_shifts}
\caption{Alignment shifts for the rotated models.}
\begin{center}
\includegraphics[width=12cm]{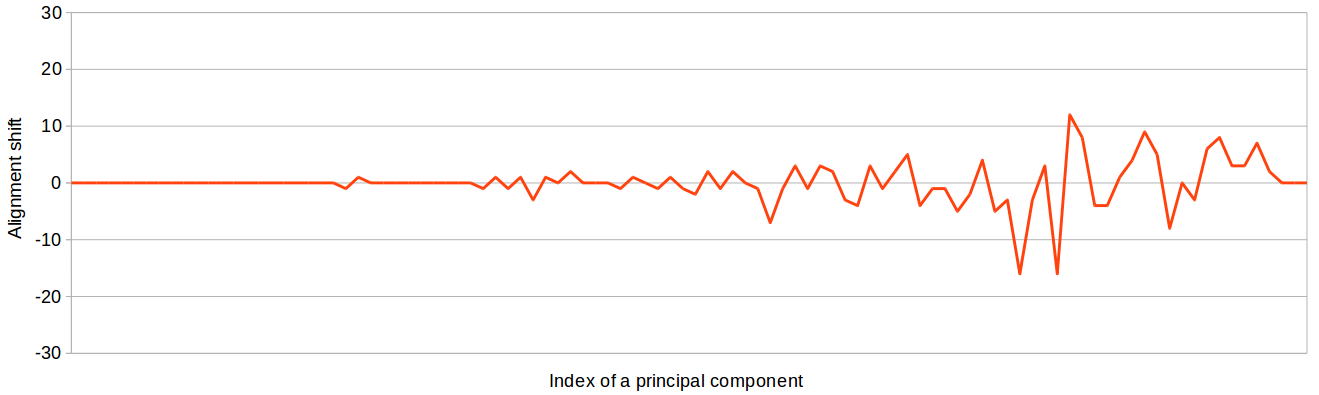}
\end{center}
\end{figure}

\begin{figure}[!t]
\caption{Normalized interpretability values for different components calculated on top/bottom 50 words for each component in source coordinates (blue) and principal coordinates (red).}
\begin{center}
\includegraphics[width=12cm]{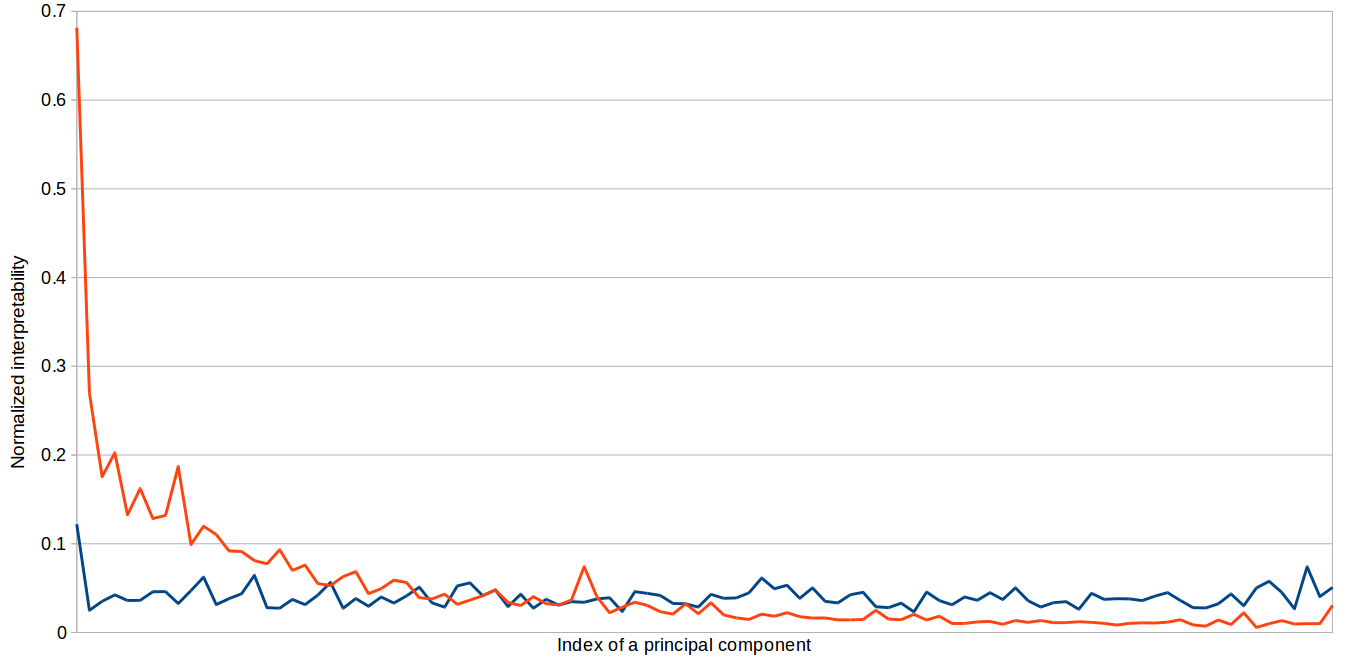}
\end{center}
\end{figure}

\subsection{Canonical basis for embeddings}
We train two fastText skipgram models on the Russian Wikipedia with default parameters.
First, we normalize all word vectors.
Then we build SVD decompositions\footnote{With numpy.linalg.svd it took up to several minutes for 100K vocabulary.} of obtained word matrices and use $V$ as an orthogonal transformation.
Thus, new ``rotated'' word vectors are described by the matrix $WV = U \Sigma$.
The corresponding singular values are shown in Figure~1, they almost coincide for both models (and thus are shown only for the one model). 
For each component both in the source and the rotated models we take top 50 words with maximal (positive)
and bottom 50 words with minimal (negative) values of the component. Taking into account that principal components are determined up to the direction,
we join these positive and negative sets together for each component.

We measure the overlapping of these sets of words. Additionally, we use the following alignment of components:
first, we look for the free indices $i$ and $j$ such that $i$th set of words from the first model and $j$th set of words from the second model
have the maximal intersection, and so on. We call the difference $i - j$ the alignment shift for the $i$th component. Results are presented in Figures 2 and 3. 

We see that at least for the first part of principal components (in the rotated models) the overlapping is big enough and is much larger that that for the source models. Moreover, these first components have almost zero alignment shifts. Other principal components have very similar singular values, and thus they cannot be determined uniquely with high confidence.

Normalized interpretibility measures for different components (calculated for 50 top/bottom words) for the source and the rotated models are shown in Fig.~4.

\subsection{Principal components of different models}
We took the following already published models:
\begin{itemize}
\item RusVect\=or\=es\footnote{\url{http://rusvectores.org/ru/models/}} lemmatized models (actually, word2vec) trained on different Russian corpora~\cite{KutuzovKuzmenko2017};
\item Russian Distributional Thesaurus\footnote{\url{https://nlpub.ru/Russian_Distributional_Thesaurus}} (actually, word2vec skipgram) models trained on Russian books corpus~\cite{Panchenko:17:RDT};
\item fastText\footnote{\url{https://github.com/facebookresearch/fastText/blob/master/pretrained-vectors.md}} model trained on Russian Wikipedia~\cite{bojanowski2016enriching}.
\end{itemize}
For each model we took $n = 10000$ or $n = 100000$ most frequent words.
Each word vector was normalized in order to replace cosines with dot products.
Then we perform SVD $W = U \Sigma V^T$ and take the matrix $W V = U \Sigma$.
For each of $d$ components we sort the words by its value and choose top $t$ ``positive'' and bottom $t$ ``negative'' words ($t=15$ or 30).
For clarity, every selection was clustered into buckets with the simplest greedy algorithm: list the selected words in decreasing order of frequency and either add the current word to some cluster if it is close enough to the word (say, the cosine is greater than $0.6$), or make a new cluster. The cluster's vector is the average vector of its words.
Intuitively, the smaller the number of clusters, the more interpretable the component is. Similar approach was used in~\cite{ramrakhiyani2017measuring}.


Tables in the Appendix show the top ``negative'' and ``positive'' words of the first principal components for different models.
We underline that principal components are determined up to the direction, and thus the separation into ``negative'' and ``positive'' parts is random. The full results are available at \url{https://alzobnin.github.io/}.
We cluster these words as described above; different clusters are separated by semicolons.
We see the following interesting features in the components:
\begin{itemize}
\item stop words: prepositions, conjunctions, etc. (RDT 1, fastText 1; in RusVect\=or\=es models they are absent just because they were filtered out before training);
\item foreign words with separation into languages (fastText 2, web 2), 
words with special orthography or tokens in broken encoding (not presented here);
\item names and surnames (RDT 8, fastText 3, web 3), including foreing names (fastText 9, web 6);
\item toponyms (not presented here) and toponym descriptors (web 7); 
\item fairy tale characters (fastText 6);
\item parts of speech and morphological forms (cases and numbers of nouns and adjectives, tenses of verbs);
\item capitalization (in fact, first positions in the sentences) and punctuation issues (e.~g., non-breaking spaces);
\item Wikipedia authors and words from Wikipedia discussion pages (fastText 5);
\item other different semantic categories.
\end{itemize}

We also made an attempt to describe obtained components automatically in terms of common contexts of common morphological and semantic tags using MyStem tagger and semantic markup from Russian National Corpus. Unfortunately, these descriptions are not as good as desired and thus they are not presented here.

\section{Conclusion}
We study principal components of publicly available word embedding models for the Russian language.
We see that the first principal components indeed are good interpretable.
Also, we show that these components are almost invariant under re-learning.
It will be interesting to explore the regularities in canonical components between different models (such as CBOW versus Skip-Gram, different train corpora and different languages~\cite{smith2017offline}. It is also worth to compare our intrinsic interpretability measure with human judgements.

\section{Acknowledgements}
The author is grateful to Mikhail Dektyarev, Mikhail Nokel, Anna Potapenko and Daniil Tararukhin for valuable and fruitful discussions.

\bibliography{paper}
\bibliographystyle{plain}

\begin{table*}[h]
\section*{Appendix}
\subsection*{Top/bottom words for the first few principal components for different Russian models}
\caption{RDT model, dim=100, 10K most frequent words}
\input{table_RDT_head10K.top15.full}
\end{table*}

\begin{table*}[!t]
\caption{fastText model, 100K most frequent words}
\input{table_fasttext_top15.head100K}
\end{table*}
\begin{table*}[!t]
\caption{RusVect\=or\=es web model, 100K most frequent words}
\input{table_rusvectores_web_ndim50.head100K.top15}

Note misspellings in 4a.
\end{table*}




\end{document}

%% file: table_RDT_head10K.top15.full.tex
\selectlanguage{russian}
\centering
\footnotesize
\begin{tabular}{|m{0.5cm} | p{11.5cm}|}
\hline
\begin{center}1\end{center} &
\begin{tabular}{p{11.25cm}}
\\
\hline
, не что как но то так же еще только уже даже того теперь действительно
\\
\end{tabular}\\
\hline
\begin{center}2\end{center} &
\begin{tabular}{p{11.25cm}}
деятельности отношении различных следовательно основе частности  е отдельных основных посредством рамках данного определенных значительной возникновения
\\
\hline
обернулся прошептал оглянулся тихонько испуганно позвал присел крикнула повернувшись хрипло вскрикнула обернувшись оглянулась позвала нагнулся
\\
\end{tabular}\\
\hline
\begin{center}3\end{center} &
\begin{tabular}{p{11.25cm}}
тебе могу хочу понимаю скажу скажи правду сомневаюсь считаешь считаете подумай поверь согласна согласится
\\
\hline
стены вдоль справа слева видны колонны полосы виднелись посередине высотой рядами бокам
\\
\end{tabular}\\
\hline
\begin{center}4\end{center} &
\begin{tabular}{p{11.25cm}}
приказал приказ срочно прибыл штаб отправил потребовал доложил направил распоряжение распорядился выехал
\\
\hline
любовь любви душа страсти природа страсть красоты красота красоту человеческая печаль;
человеческой плоти человеческое
\\
\end{tabular}\\
\hline
\begin{center}5\end{center} &
\begin{tabular}{p{11.25cm}}
вечер вечера кафе обеда позвонила ресторане отеле вечерам проводила утрам;
приехала отправилась ходила мамой купила
\\
\hline
меч воин меча клинок копье взмахнул;
рявкнул вскинул выкрикнул прошипел дернулся вскрикнул завопил прорычал
\\
\end{tabular}\\
\hline
\begin{center}6\end{center} &
\begin{tabular}{p{11.25cm}}
предмет предмета;
компьютер автоматически компьютера;
рассматривать модели модель анализа включает;
клиента клиент
\\
\hline
воины враги войско волки лесах;
деревню родину родные;
боялись старики;
умирать погибнуть
\\
\end{tabular}\\
\hline
\begin{center}7\end{center} &
\begin{tabular}{p{11.25cm}}
получается короче небось блин нету хрен кой;
работают умеют берут платят;
штук поменьше
\\
\hline
гнев отчаяние волнение отчаяния испытывала охватило;
покинул встретился застал покинула;
объятиях страстно
\\
\end{tabular}\\
\hline
\begin{center}8\end{center} &
\begin{tabular}{p{11.25cm}}
николай петр павел иванович михаил василий григорий васильевич михайлович георгий федорович
\\
\hline
сможет смогу смогла готова шанс попыталась способна пытаться;
выбраться вырваться выжить сопротивляться убежать сбежать
\\
\end{tabular}\\
\hline
\begin{center}9\end{center} &
\begin{tabular}{p{11.25cm}}
опять иван ваня алеша;
начинается открывается следующая;
москва петербург киев;
весна осень
\\
\hline
мужчины мужчин;
воины эльфы;
казались выглядели представляли напоминали являлись отличались позволяли держались
\\
\end{tabular}\\
\hline
\begin{center}10\end{center} &
\begin{tabular}{p{11.25cm}}
хлеб хлеба;
посуду ложку
\\
\hline
видел встречались;
произошло творится нахожусь;
находится существует знаем;
планета станция
\\
\end{tabular}\\
\hline
\end{tabular}

%% file: table_fasttext_top15.head100K.tex
\selectlanguage{russian}
\centering
\footnotesize
\begin{tabular}{|m{0.5cm} | p{11.5cm}|}
\hline
\begin{center}1\end{center} &
\begin{tabular}{p{11.25cm}}
\\
\hline
, . и а как во же том того пор репосты/рапорты/проверенные бессвязное» репосты/рапорты;
взрываемости кмет\#болгариякмет
\\
\end{tabular}\\
\hline
\begin{center}2\end{center} &
\begin{tabular}{p{11.25cm}}
царской царского царских царским мещан велено округи надлежало считаясь ходатайствовать деятельно петровских
\\
\hline
mr tom another third chris joe eric alone larry presents ron singer jennifer trailer alternate
\\
\end{tabular}\\
\hline
\begin{center}3\end{center} &
\begin{tabular}{p{11.25cm}}
богданович михайло бельский данило христо петро емельян василь рыльский гришко калиш назарий конюх владимир  любин
\\
\hline
позволяет использовании учитывать определять отличаться целесообразно зависеть функциональности изменяться различаться упрощает минимизировать приемлемой потребоваться оптимизировать
\\
\end{tabular}\\
\hline
\begin{center}4\end{center} &
\begin{tabular}{p{11.25cm}}
хотел сказав убеждает простить восторге поверил соблазнить разочарован простил ненавидел обманул отговорить сожалеет рассказав проникся
\\
\hline
магистральных котельных лесхоз сортировочный подстанций мелиоративных нижнегорский камско торфопредприятия кировско вагонное тракторных;
серебряно дерново казанка
\\
\end{tabular}\\
\hline
\begin{center}5\end{center} &
\begin{tabular}{p{11.25cm}}
оконечности сантиметров суше передвигаться льдом укрытия воздуху передвигается спускаются передвигаются канаты повредив сбросив стволами перемещаясь
\\
\hline
авторитетность обращаю читаем цитирую mitrius волохонский kak thejurist jannikol пиотровский критика» авторитетно сомневаетесь fhmrussia chelovechek
\\
\end{tabular}\\
\hline
\begin{center}6\end{center} &
\begin{tabular}{p{11.25cm}}
заяц мужик старуха шарик нежный нежно шапочка бледный очи мышонок глазки солнышко ёжик леший старухи
\\
\hline
правительством соглашения объявило предоставлении соглашением конгрессом подписанием финансировании реструктуризации предоставило подписало директорат подписанию соглашениям финансированию
\\
\end{tabular}\\
\hline
\begin{center}7\end{center} &
\begin{tabular}{p{11.25cm}}
машину водитель авто авиа автомобилист дублёр отработал рекорд» стажёр отработать кц подключился;
площадке старт» чп
\\
\hline
xiixiii xiii в xixii xiiixiv xii в iii в ii в xi в vii в xxi vvi viii в viiiix vivii x в
\\
\end{tabular}\\
\hline
\begin{center}8\end{center} &
\begin{tabular}{p{11.25cm}}
творчества художественной художественного классической творческой классических творческого музыки» пластической искусства  пластических исполнительского фортепианной кинематографического пластического
\\
\hline
блокировать заблокировать заблокировал воевать откатывать патрулировать заблокированы заблокировали блокировали вешать блокировал блокирован вандалить откатили удалит
\\
\end{tabular}\\
\hline
\begin{center}9\end{center} &
\begin{tabular}{p{11.25cm}}
выпущены издавалась ставились исполнялась продавались исполнены визитной исполнялись украшали открывали демонстрировались выходившая выходившие открывала
\\
\hline
джефферсон чавес вильсон луа очоа барре прието макартур арсе мугабе салазар ходж друз зума
\\
\end{tabular}\\
\hline
\begin{center}10\end{center} &
\begin{tabular}{p{11.25cm}}
ум адъютантом действительного приходился смещён последователем  служившего ординарного  сообщ смещен non\_performing\_personnel
\\
\hline
местные национальные регионы региональные азиатские рестораны туристические развлекательные корейские тигры аборигены бары миллионеры мигранты индонезийские
\\
\end{tabular}\\
\hline
\end{tabular}

%% file: table_rusvectores_web_ndim50.head100K.top15.tex
\selectlanguage{russian}
\centering
\footnotesize
\begin{tabular}{|m{0.5cm} | p{11.5cm}|}
\hline
\begin{center}1\end{center} &
\begin{tabular}{p{11.25cm}}
информация{\tiny noun} услуга{\tiny noun} предложение{\tiny noun} оплата{\tiny noun} получение{\tiny noun} законодательство{\tiny noun} размещение{\tiny noun} работодатель{\tiny noun} заинтересованный{\tiny adj} трудоустройство{\tiny noun} соискатель{\tiny noun};
условие{\tiny noun} необходимый{\tiny adj} независимо{\tiny adv}
\\
\hline
анонсирование::плэйкастовый{\tiny noun} непросмотренный::резюме{\tiny noun} tools::trade{\tiny noun} webkind{\tiny noun} support::infon{\tiny noun} yellcity{\tiny noun} elec::elec{\tiny noun} spell::correction{\tiny noun} миколь::гоголь{\tiny noun} ненормовані{\tiny noun} электроника::techhome{\tiny noun} copyright::restate{\tiny noun} fannet::org{\tiny noun} своб::индексир{\tiny noun} ted::lapidus{\tiny noun}
\\
\end{tabular}\\
\hline
\begin{center}2\end{center} &
\begin{tabular}{p{11.25cm}}
value{\tiny noun} plus{\tiny noun} classic{\tiny noun} super{\tiny noun} light{\tiny noun} series{\tiny noun} tech{\tiny noun} standard{\tiny noun} horizon{\tiny noun} cyber{\tiny noun} regular{\tiny noun} circuit{\tiny noun} isis{\tiny noun};
blue{\tiny noun} gold{\tiny noun}
\\
\hline
сказать{\tiny verb} знать{\tiny verb} говорить{\tiny verb} приходить{\tiny verb} спрашивать{\tiny verb} пойти{\tiny verb} подумать{\tiny verb} решаться{\tiny verb} удивляться{\tiny verb} припоминать{\tiny verb} впрямь{\tiny adv} недоумевать{\tiny verb} сговариваться{\tiny verb} отчего-то{\tiny adv} помалкивать{\tiny verb}
\\
\end{tabular}\\
\hline
\begin{center}3\end{center} &
\begin{tabular}{p{11.25cm}}
сделать{\tiny verb} делать{\tiny verb} забывать{\tiny verb} просто{\tiny adv} угодно{\tiny part} надоедать{\tiny verb} любой{\tiny pron} пугаться{\tiny verb} чертовски{\tiny adv} may::captain{\tiny noun} черт::побрать{\tiny verb};
смотреть{\tiny verb} посмотреть{\tiny verb}
\\
\hline
попов{\tiny noun} андреев{\tiny noun} калинин{\tiny noun} максимов{\tiny noun} мельников{\tiny noun} тихомиров{\tiny noun} емельянов{\tiny noun} кондратьев{\tiny noun} румянцев{\tiny noun} романовский{\tiny noun} андрианов{\tiny noun} чеботарев{\tiny noun} горячев{\tiny noun} моисеенко{\tiny noun} чудновский{\tiny noun}
\\
\end{tabular}\\
\hline
\begin{center}4
\end{center} &
\begin{tabular}{p{11.25cm}}
огнетушитель{\tiny noun} балоны{\tiny noun} балон{\tiny noun} закрутка{\tiny noun} грузик{\tiny noun} обьем{\tiny noun} приспособ{\tiny noun} воздушка{\tiny noun} акуратно{\tiny adv} собраный{\tiny adj};
железка{\tiny noun} ессный{\tiny adj}
\\
\hline
британский{\tiny adj} роберт{\tiny noun} джордж{\tiny noun} известность{\tiny noun} влиятельный{\tiny adj} габриэль{\tiny noun} фрэнсис{\tiny noun} коэн{\tiny noun} чарлз{\tiny noun} гарольд{\tiny noun} эдмунд{\tiny noun} фредерика{\tiny noun} тэтчер{\tiny noun} теодора{\tiny noun} джулиана{\tiny noun}
\\
\end{tabular}\\
\hline
\begin{center}5\end{center} &
\begin{tabular}{p{11.25cm}}
образ{\tiny noun} лишь{\tiny part} род{\tiny noun} человеческий{\tiny adj} глубокий{\tiny adj} прежде{\tiny adp} естественный{\tiny adj} характерный{\tiny adj} подобно{\tiny adp} по-видимому{\tiny adv} отчетливый{\tiny adj} рые{\tiny noun} рый{\tiny noun} редуцированный{\tiny adj}
\\
\hline
позвонить{\tiny verb} звонить{\tiny verb};
присылать{\tiny verb} отписываться{\tiny verb};
привет{\tiny noun} личка{\tiny noun} зарегестрировать{\tiny verb} зарегистрированный{\tiny adj}
\\
\end{tabular}\\
\hline
\begin{center}6\end{center} &
\begin{tabular}{p{11.25cm}}
чаяние{\tiny noun} суетный{\tiny adj} обличение{\tiny noun} безбожный{\tiny adj} своеволие{\tiny noun} леность{\tiny noun} властолюбие{\tiny noun} чуждаться{\tiny verb} иоаннов{\tiny noun} вековечный{\tiny adj} юродство{\tiny noun}
\\
\hline
пит{\tiny noun} брэд{\tiny noun} бишоп{\tiny noun} пирсон{\tiny noun} куин{\tiny noun} филд{\tiny noun} мосс{\tiny noun} кроуфорд{\tiny noun} дафф{\tiny noun} уолтерс{\tiny noun} дэйли{\tiny noun} слоун{\tiny noun} роуч{\tiny noun} макинтайра{\tiny noun}
\\
\end{tabular}\\
\hline
\begin{center}7\end{center} &
\begin{tabular}{p{11.25cm}}
город{\tiny noun} дом{\tiny noun} улица{\tiny noun} парк{\tiny noun} столица{\tiny noun} дворец{\tiny noun} городок{\tiny noun} недалеко{\tiny adv} холм{\tiny noun} неподалеку{\tiny adv} пригород{\tiny noun} близ{\tiny adp} окрестности{\tiny noun} подножие{\tiny noun}
\\
\hline
адекватный{\tiny adj} адекватно{\tiny adv} неадекватный{\tiny adj} адекватность{\tiny noun} априори{\tiny adv} латентный{\tiny adj} неадекватность{\tiny noun} нормальность{\tiny noun} лакмусовый::бумажка{\tiny noun} когнитивный::диссонанс{\tiny noun}
\\
\end{tabular}\\
\hline
\end{tabular}